\documentclass[final,1p,times]{elsarticle}
\usepackage{amsmath,amssymb,amsfonts}
\usepackage{algorithmic}
\usepackage{graphicx,color}
\usepackage{textcomp}
\usepackage[dvipsnames]{xcolor}
\usepackage{hyperref}
\usepackage{algorithm,algorithmic}
\usepackage{multirow}
\usepackage{placeins}
\usepackage{cleveref}
\usepackage{enumitem,comment}
\usepackage{booktabs}

\usepackage{fancyhdr}
\usepackage{hyperref}

\usepackage[inkscapelatex=false]{svg}

\makeatletter

\begin{document}

\begin{frontmatter}

\title{Squeeze-Release: Iterative Pruning with Exact Structural Minimization}


\author[label1]{Roman Denkin\corref{cor1}}
\ead{roman.denkin@it.uu.se}

\author[label1]{Ida Åkerholm}

\author[label1,label2]{Prashant Singh}

\author[label1,label2]{Ida-Maria Sintorn}

\cortext[cor1]{Corresponding author.}

\affiliation[label1]{organization={Department of Information Technology, Uppsala University},
             city={Uppsala},
             country={Sweden}}

\affiliation[label2]{organization={Science for Life Laboratory, Uppsala University},
             city={Uppsala},
             country={Sweden}}

\begin{abstract}
Unstructured pruning produces sparse weight tensors, but the standard implementation keeps tensor shapes unchanged so the deployed model is no smaller than before pruning. 
We present an exact structural rewrite, which we call minimization, that converts a masked network into a smaller dense network with the same forward function up to floating-point rounding. 
The Squeeze-Release cycle iterates pruning and minimization with an intermediate release step that re-enables the exact-zero positions inside the compacted tensors as small calibrated noise, turning otherwise wasted capacity back into trainable parameters. 
Successive cycles use that capacity to find structural redundancy a single pass cannot reach. 
We additionally introduce CompensatedLayerNorm, a function-preserving replacement for LayerNorm that extends minimization to channel reduction across LayerNorm-equipped residual streams. 
Squeeze-Release compresses the deployable network to $39\times$ smaller than the unpruned model on a fully-connected model network and $14.8\times$ smaller on modern CNN (ConvNeXt-Tiny), at comparable accuracy. 
In addition we prove that the rewrite can be extended to transformer architectures.
\end{abstract}


\end{frontmatter}

\section{Introduction}
\label{sec:intro}

The purpose of network pruning is to remove parameters from a trained model so it uses less memory and compute when deployed (at inference).
Unstructured pruning, which zeros out individual weights according to an importance criterion, reaches high sparsity at a small accuracy cost and is widely used \cite{hoefler2021sparsity}.
What such a pruner returns, however, is a set of weight tensors in which most entries are zero and the tensor shapes are unchanged.
PyTorch's \texttt{torch.nn.utils.prune}, the standard reference implementation, stores both the original parameter \texttt{weight\_orig} and a boolean \texttt{weight\_mask} and applies the mask as a forward-time multiplication; the call \texttt{prune.remove} folds the mask into the parameter but does not change any tensor shape.
The on-disk/in-memory model is therefore at least as large as the unpruned model, and every forward pass still runs the full dense kernel.
The count of mask-alive parameters, the metric most pruning papers report, accordingly does not correspond to the size of any runnable artifact, and it overstates what general-purpose hardware can deliver: speedups from unstructured sparsity on commodity GPUs are typically small \cite{mishra2021accelerating, elsen2020fastsparse}.
The deployable size, meaning the smallest dense network with the same forward function, is what determines memory, latency, and runtime cost on standard hardware, and the gap between it and the mask-alive count can be wide because unstructured pruning rarely produces the channel-aligned or neuron-aligned zero patterns required to physically narrow a layer.

Pruning involves two distinct problems: deciding which parameters to remove, and actually removing them.
The first concerns importance scoring, training schedules, and fine-tuning protocols, and accounts for most of the published work.
The second is a structural-rewrite problem: given a network with a known sparsity pattern, produce a smaller dense network with the same forward function.
This paper focuses on the second problem.
We call its solution \emph{minimization}: an exact rewrite that converts a masked or pruned network into a smaller dense network whose output matches the original up to floating-point rounding.
Minimization is independent of how the sparsity pattern was obtained, so it applies as a one-shot post-processing step on top of any unstructured pruner.
It also serves as the inner transformation of an outer iterative loop.

Even after minimization, the resulting dense network is not maximally compact.
Pruning at the level of individual weights drives many entries to zero, but only a subset of those zeros aligns into the whole-channel or whole-neuron patterns that minimization can structurally remove.
The rest persist inside the kept dense tensors, where they are stored, loaded, and multiplied on every forward pass without contributing to the output.
This is \emph{wasted computational capacity} in the deployable network.
Re-enabling those positions as trainable parameters returns them to the optimization at no size or compute cost.
Applying this re-enabling step in alternation with further pruning gives the network repeated opportunities to consolidate its useful computation capacity.
Whether such a cyclic procedure produces a smaller deployable network than a single prune-and-minimize pass is one of the questions this paper addresses.

Our main contribution is \emph{Squeeze-Release}, an iterative procedure in which each cycle prunes the current network, minimizes it (\emph{squeeze}), replaces the previously disabled positions in the squeezed tensors with small random values calibrated to the surviving layer statistics (\emph{release}), and fine-tunes the released model.
The cycle uses minimization as its inner transformation, with the goal of reaching smaller deployable networks than a single pass of pruning and minimization, while using all available computational power in its smaller shape.
Our second contribution is \emph{CompensatedLayerNorm}, a function-preserving replacement for \texttt{LayerNorm} that uses three scalar statistics of the removed channels (count, sum, sum-of-squares) which is sufficient to reconstruct full-width normalization exactly.
LayerNorm is widely used in modern architectures, and reducing the channel dimension across a layer in a function-preserving way is non-trivial; to the best of our knowledge, prior pruning work has avoided or approximated this case rather than performing the reduction exactly.

\section{Related Work}
\label{sec:related}

The idea of pruning trained networks by ranking parameters with an importance score dates back to Optimal Brain Damage \cite{lecun1989obd} and Optimal Brain Surgeon \cite{hassibi1992obs}, which used second-derivative information to identify removable weights.
The modern deep-network pruning revival comes from Han et al~\cite{han2015learning} with magnitude-based ranking; Frankle et al~\cite{frankle2019lottery} reframed the picture through the lottery-ticket hypothesis, and Hoefler et al~\cite{hoefler2021sparsity} survey the broader literature.
Structured pruning, which removes whole channels or filters \cite{wen2016ssl}, sidesteps the deployment gap by construction; at matched compression level, however, unstructured pruning (removing individual weights and biases) generally upper-bounds the accuracy achievable by structured methods \cite{gale2019state}.
On the unstructured side, sparse kernels for arbitrary patterns rarely match dense throughput on commodity GPUs, so the predicted savings from a pruning ratio do not, on their own, translate into deployment-time gains \cite{mishra2021accelerating, elsen2020fastsparse, gale2020sparse}. Effort has been put into accelerating unstructured sparsity, especially by developing custom hardware accelerators \cite{zhang2021snap, gondimalla2023eureka}, but methods achieving speedups on commodity GPUs are missing. Minimization is the structural-rewrite step that closes this gap for any unstructured pruner, and it is useful even as a one-shot post-processing step on top of an existing pipeline.

Gradual pruning schedules \cite{zhu2017prune, zhu2025fggp} alternate between pruning and fine-tuning steps to allow the model to recover from pruning damage.
A separate line re-enables pruned weights during dense retraining phases, including Dense-Sparse-Dense training \cite{han2017dsd} and AC/DC \cite{peste2021ac}.
Both styles operate on a fixed-shape network throughout.
A different line iteratively shrinks the network itself: multi-pass Network Slimming \cite{liu2017slimming} prunes channels and rebuilds a narrower model across passes, Minitron \cite{muralidharan2024minitron} chains prune-and-distill rounds, and PruneTrain \cite{lym2019prunetrain} reconfigures the network periodically during training while explicitly arguing that pruned weights "almost never revive."
None of these iterates while both physically shrinking the network and re-enabling pruned positions between rounds.

In a LayerNorm-equipped residual stream the normalization statistics are computed across the channel dimension, so removing channels changes the per-token mean and variance and therefore the output.
Most ViT and LLM pruning work avoids the problem by pruning only the attention-internal and MLP-internal dimensions \cite{ma2023llmpruner, chavan2022vitslim}.
Methods that do reduce residual width typically rely on long fine-tuning to recover and do not use any special methods to preserve exact LayerNorm output\cite{yang2023nvit, xia2024shearedllama}.
Two recent methods engage with the issue directly but sidestep exact preservation: SliceGPT \cite{ashkboos2024slicegpt} converts LayerNorm to RMSNorm and uses orthogonal-rotation invariance to slice the rotated representation, and Pangu Light \cite{chen2025pangulight} rescales the affine parameters of RMSNorm after pruning as a stabilization heuristic.
To the best of our knowledge, no published method reconstructs the LayerNorm statistics exactly from sufficient statistics of the removed channels, which is what makes the channel reduction in CompensatedLayerNorm function-preserving.

Function-preserving transformations of neural networks are operations that change the architecture while keeping the input-output function intact.
The category is overwhelmingly aimed at growing networks: Net2Net \cite{chen2016net2net} introduces wider and deeper layers, Network Morphism \cite{wei2016morphism} generalizes to a richer set of architectural changes, GradMax \cite{evci2022gradmax} initializes added neurons by maximizing their gradient norm, and recent work extends the framework to residual connections \cite{painter2024fpt}.
The shrinking direction is sparsely covered: Neuron Merging \cite{kim2020merging} compensates for pruned neurons by combining them with similar surviving ones via a cosine-similarity decomposition, but the construction is approximate and only applies under ReLU.
Our minimization is the exact shrinking dual: it removes neurons, channels, and dead residual blocks while reproducing the original forward function up to floating-point rounding.

\section{Method}
\label{sec:method}

\subsection{Problem setting and notation}
\label{sec:method-setup}

Let $\theta$ denote the trainable parameters of a neural network - in this work, the weights of \texttt{Linear} and \texttt{Conv2d} layers; biases and normalization affine parameters are not pruned. Unstructured pruning attaches a binary mask $M \in \{0,1\}^{|\theta|}$ and replaces each weight tensor with the elementwise product $\theta \odot M$ at forward time, while keeping $\theta$ itself unchanged. 
Throughout the paper we report two parameter counts. 
The \textbf{mask-alive} count $\|M\|_0$ is what most unstructured-pruning work reports as ``the pruned size''. The \textbf{minimal} count is the parameter count of the dense network produced by the structural rewrite of \cref{sec:method-minimize}, i.e.\ the size of an actually deployable artifact. 
The minimal count is at most the mask-alive count plus all weights in rows, columns and filters that the rewrite cannot collapse; in practice \textbf{mask-alive} count and \textbf{minimal} count diverge substantially, and this gap is quantified by our experiments in~\cref{sec:experiments}.

A pruning step assigns a scalar importance score to each prunable unit and removes the lowest-scoring units up to a target sparsity. We use this in the standard iterative form: prune, fine-tune on the training data, repeat.

To decide which parameters to remove, a scoring function is needed to rank the parameters. 
Common scores include Hessian-based scores \cite{lecun1989obd, hassibi1992obs}, weight magnitude \cite{han2015learning} and gradient-based scores \cite{lee2018snip, tanaka2020synflow}. 
The pruning score used in our experiments is the product of gradient and weight magnitude $|\nabla_\theta L \cdot \theta|$, calculated on the last batch of training data before pruning. 
This pruning score has previously been shown useful in both structural iterative pruning \cite{molchanov2017pruning}, and for unstructured one-shot pruning \cite{lee2018snip, tanaka2020synflow}. In \cite{tanaka2020synflow}, a general class of scores named synaptic saliencies is defined. 
This class includes all scores on the form $s(\theta) = \frac{\delta R}{\delta \theta} * \theta$, where R is a scalar loss function. 

The granularity at which scores are assigned and weights are removed is dictated by what minimization can later collapse. 
For a \texttt{Linear} layer, scalar pruning is sufficient: a row of the weight matrix that becomes entirely zero leaves the corresponding output neuron with an input-independent post-activation, and a column that becomes entirely zero indicates an unread input; both cases are removed by the rewrite of \cref{sec:method-minimize}. For a \texttt{Conv2d} layer, scalar pruning does not enable structural collapse: zeroing arbitrary entries of a weight tensor of shape $(C_\text{out}, C_\text{in}, k, k)$ leaves all four dimensions unchanged, and the only unit that minimization can remove is an entire output filter - a slice $\theta[j, :, :, :]$ that is fully zero. 
We therefore score and remove \texttt{Conv2d} weights at \emph{filter} granularity: each output filter receives a single score, obtained by averaging the per-element scores within the filter, and is either kept or removed as a whole. 
\texttt{Linear} layers in the same network (the point-wise convolutions inside ConvNeXt blocks, the classifier head) remain element-wise. 
The two granularities are pooled into a single global selection so that filters and individual \texttt{Linear} weights compete on the same score scale, with each filter accounting for as many ``slots'' as it contains scalar weights.

\subsection{Exact minimization (\emph{squeeze}): dead-incoming and dead-outgoing units}
\label{sec:method-minimize}

Minimization rewrites a masked network as a smaller dense network with the same forward function up to floating-point rounding. 
It removes the structural redundancy that the binary mask introduces, replacing each masked weight tensor with a smaller dense tensor and discarding the mask buffers altogether. 
The rewrite operates on whole units of the network: neurons of \texttt{Linear} layers and output filters of \texttt{Conv2d} layers.

\subsubsection{Fully-connected networks}
\label{sec:method-fc}

Consider a layer $i$ being minimized, with weight $W_i$ and bias $b_i$. Its outputs pass through an activation $\phi$ before reaching the consumer layer $i{+}1$ with weight $W_{i+1}$ and bias $b_{i+1}$. 
We call a unit $k$ of layer $i$ \emph{dead-incoming} if row $k$ of $W_i$ is entirely zero. 
In that case the pre-activation of $k$ is determined by $b_i[k]$ alone, so its post-activation output is the input-independent constant $c_k = \phi(b_i[k])$. 
A neuron-wise normalization between the linear layer and the activation, such as the BatchNorm used in our FC model and evaluated in inference mode, does not affect this argument: it acts elementwise on $b_i[k]$ and produces another constant, which we absorb into $c_k$. We call $k$ \emph{dead-outgoing} if column $k$ of $W_{i+1}$ is entirely zero, so the consumer does not read $k$.

\begin{figure}[t]
  \centering
  \includegraphics[width=\linewidth]{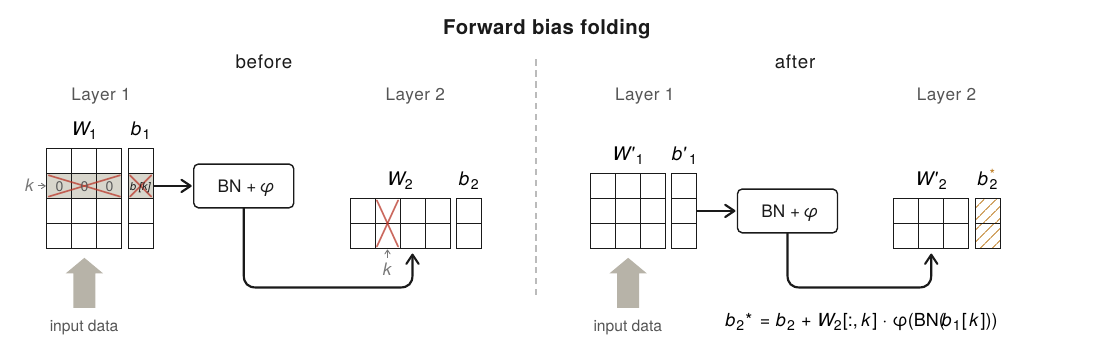}
  \caption{Forward bias folding for a dead-incoming neuron. When row $k$ of $W_1$ is structurally zero, the neuron $k$'s post-activation output is the constant $c_k = \varphi(\mathrm{BN}(b_1[k]))$. Removing neuron $k$ without changing the network's input--output map requires adding $W_2[:,k]\,c_k$ into $b_2$, after which row $k$ of $W_1$ and column $k$ of $W_2$ can be dropped.}
  \label{fig:bias-folding}
\end{figure}

Each condition admits an exact structural rewrite. 
For a dead-incoming unit, the consumer's contribution from $k$ is the fixed vector $W_{i+1}[:, k]\,c_k$, which we absorb into the consumer's bias as $b_{i+1} \leftarrow b_{i+1} + W_{i+1}[:, k]\,c_k$, and then drop row $k$ of $W_i$ together with column $k$ of $W_{i+1}$. 
The operation preserves the forward function exactly because the consumer is linear in its inputs. \Cref{fig:bias-folding} illustrates this rewrite on a toy two-layer stack. 
For a dead-outgoing unit, the output of $k$ never enters any downstream computation, so we drop row $k$ of $W_i$ together with the zero column $k$ of $W_{i+1}$. 
When both conditions hold for the same $k$, the dead-outgoing removal alone is sufficient: the consumer does not read $k$, so a dead-incoming fold would multiply $c_k$ by the zero column of $W_{i+1}$ and contribute nothing.

We apply this rewrite to each linear layer of the network in turn. 
A constant produced by folding at layer $i$ becomes a contribution to $b_{i+1}$; if a row of $W_{i+1}$ is itself all-zero, the corresponding unit is dead-incoming with respect to the updated bias, and the next iteration of the rewrite absorbs its constant output into layer $i+2$. 
Cascading constants are therefore handled implicitly, without a separate pass. 
The dead-outgoing case has a small variation at the input. 
An input coordinate not read by any weight in the first layer can be dropped like any other dead-outgoing unit, except that there is no upstream producer whose row needs removing; we record the unused coordinates in a persistent input mask applied before the first matrix multiplication, and drop the corresponding columns of the first layer's weight.

The dead-incoming fold has a narrow precedent, where a BatchNorm channel trained to $\gamma = 0$ followed by ReLU is folded into the next layer in the same form\cite{ye2018rethinking}. 
The rewrite presented here covers arbitrary dead-incoming units, regardless of how the all-zero row arose, and any elementwise activation.

\subsubsection{ConvNeXt}
\label{sec:method-convnext}

ConvNeXt \cite{liu2022convnet} is a convolutional image-classification architecture that adopts several design choices from vision transformers while keeping a hierarchical CNN structure. 
A ConvNeXt model consists of a convolutional stem followed by four stages of identical-shape blocks, with a downsampling layer between stages and a linear classifier on top. 
Each stage operates on a fixed channel width $C$, and within a stage the residual stream carrying the $C$-wide block input is preserved across blocks. 
A block is an inverted-bottleneck unit, i.e., a depthwise $7{\times}7$ convolution (DWConv) followed by LayerNorm, then two pointwise convolutions PWConv1 and PWConv2 expanding the channel width from $C$ to $4C$ and back, with GELU between them.
The branch output is added to the residual.

\begin{figure}[t]
\centering
\includegraphics[width=\linewidth]{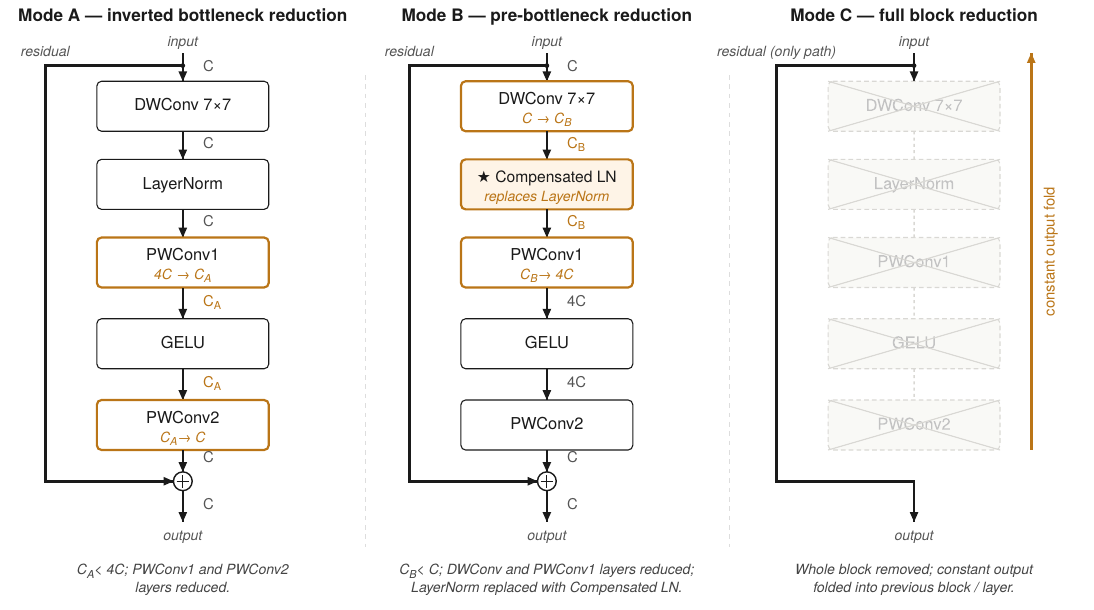}
\caption{ConvNeXt block under the three minimization modes. \textbf{Mode~A} (inverted bottleneck reduction) reduces the inner $4C$ dimension to $C_A < 4C$ by modifying \texttt{PWConv1} and \texttt{PWConv2}. \textbf{Mode~B} (pre-bottleneck reduction) reduces the pre-bottleneck width to
$C_B < C$ by modifying \texttt{DWConv} and \texttt{PWConv1}; \texttt{LayerNorm} is replaced with \emph{Compensated LayerNorm}, which reproduces the full-width LN output from stored sufficient statistics of the removed constant channels (see~\Cref{sec:method-cln}). \textbf{Mode~C} (full block reduction) removes a block which does not consume any input anymore; the resulting constant output is folded backward into the predecessor's bias (stem \texttt{LN}, downsample conv, or previous block's \texttt{PWConv2}).
Amber color mark layers modified relative to the unreduced block.}
\label{fig:convnext-block}
\end{figure}

In the implementation we work with, PWConv1 and PWConv2 are realised as \texttt{Linear} layers operating on the channel dimension after a spatial permute, while DWConv is a \texttt{Conv2d}. 
The corresponding pruning units (similar to fully-connected networks in \cref{sec:method-setup}) are individual scalar weights for the pointwise layers and entire filters for DWConv. 
The dead-incoming and dead-outgoing analysis of \cref{sec:method-fc} therefore applies directly to PWConv1 and PWConv2; for DWConv it applies at filter granularity, where ``row $k$ is zero'' means the spatial kernel producing output channel $k$ is entirely zero. 
Three minimization modes act on different parts of the blockas illustrated in~\cref{fig:convnext-block}.

The first mode reduces the $4C$-wide intermediate between PWConv1 and PWConv2 (\cref{fig:convnext-block}, Mode~A). 
Between these two layers the only operation is GELU, with no normalization, so this mode is a direct application of the FC rewrite to the two pointwise layers. 
A channel $k$ of the $4C$ width is dead-incoming if row $k$ of PWConv1's weight is entirely zero, in which case its post-GELU output is the constant $c_k = \mathrm{GELU}(b_{\text{pw1}}[k])$; it is dead-outgoing if column $k$ of PWConv2's weight is entirely zero. 
The rewrites match \Cref{sec:method-fc}: for a dead-incoming channel, fold $W_{\text{pw2}}[:, k]\,c_k$ into PWConv2's bias and then drop the corresponding row of PWConv1 together with column $k$ of PWConv2; for a dead-outgoing channel, drop the same row and column without folding.

The second mode reduces the inner channel width of the path from DWConv through LayerNorm to PWConv1 (\cref{fig:convnext-block}, Mode~B). 
The complication here is LayerNorm. 
BatchNorm normalizes each channel independently using the channel's own running statistics, so removing a channel from its input does not affect any surviving channel's output, and the rewrite of \cref{sec:method-fc} carries over unchanged. 
LayerNorm instead computes the mean and variance across channels at every spatial position, so removing a channel changes the statistics that LayerNorm assigns to every surviving channel at that position, and the dead-incoming or dead-outgoing condition alone is not enough for an exact rewrite. 
We can still remove a channel $k$ when both conditions hold simultaneously: the DWConv filter producing channel $k$ is entirely zero, so after DWConv the channel is the spatial constant $b_{\text{dw}}[k]$, and PWConv1's column reading channel $k$ is entirely zero. 
Under these conditions the channel contributes an offset to LayerNorm's per-position statistics that does not depend on the input, and a modified LayerNorm that we call CompensatedLayerNorm absorbs this offset's effect on the surviving channels' normalization; its construction is given later. 
The DWConv filter for channel $k$ and the corresponding column of PWConv1 can then be dropped.

The third mode removes an entire block (\cref{fig:convnext-block}, Mode~C). 
When every DWConv filter of a block is entirely zero, the input to LayerNorm is a per-channel spatial constant and the branch output evaluates to a fixed per-channel constant, so the block reduces to a fixed per-channel addition to the residual stream. 
We absorb this addition into the bias of the layer immediately upstream of the block (the previous block's PWConv2, or the stage's stem layer for the first block of a stage), after which the block is removed entirely. 
This mode triggers only in later stages of heavy pruning, once all of a block's depthwise filters have been zeroed.

\subsection{CompensatedLayerNorm: exact channel reduction across LayerNorm}
\label{sec:method-cln}

In the Mode B rewrite of \Cref{sec:method-convnext}, a channel removed from LayerNorm's input is one that has been forced to a spatial constant by the conditions on DWConv and PWConv1. 
Even though such a channel does not depend on the input, it enters the per-position mean and variance that LayerNorm computes across the channel dimension; dropping it and running standard LayerNorm on a narrower input would change the statistics seen by every surviving channel. 
CompensatedLayerNorm replaces the standard module with one that reproduces the original full-width LayerNorm output on the surviving channels exactly.

For input $x \in \mathbb{R}^C$ at one spatial position, standard LayerNorm computes
\begin{equation*}
\mu = \frac{1}{C}\sum_{i=1}^{C} x_i, \qquad
\sigma^2 = \frac{1}{C}\sum_{i=1}^{C}(x_i - \mu)^2,
\end{equation*}
\begin{equation*}
\mathrm{LN}(x)_i = \gamma_i\,\frac{x_i - \mu}{\sqrt{\sigma^2 + \varepsilon}} + \beta_i.
\end{equation*}
Now, partition the $C$ channel indices into the kept set $A$ with $|A| = m$ and the removed set $R$ with $|R| = K = C - m$, where each $j \in R$ is the spatial constant $c_j$.
CompensatedLayerNorm stores the affine parameters $\gamma, \beta$ restricted to $A$ together with three scalar summaries of $R$: the count $K$, the sum $S = \sum_{j \in R} c_j$, and the sum of squares $Q = \sum_{j \in R} c_j^2$.
In a forward pass it receives only the $m$ kept inputs, computes their mean $\mu_A$ and second central moment $\sigma_A^2$, and reconstructs the full-width $\mu$ and $\sigma^2$ above.

Both $\mu$ and $\sigma^2$ are sums over all $C$ indices and split by the index set:
\begin{equation*}
C\mu = m\,\mu_A + S, \qquad
C\sigma^2 = \sum_{i \in A}(x_i - \mu)^2 + \sum_{j \in R}(c_j - \mu)^2.
\end{equation*}
The first identity gives $\mu$ directly.
For the second, expand $(x_i - \mu) = (x_i - \mu_A) + (\mu_A - \mu)$ inside the kept-channel sum and square:
\begin{align*}
\sum_{i \in A}(x_i - \mu)^2
&= \sum_{i \in A}(x_i - \mu_A)^2 + 2(\mu_A - \mu)\sum_{i \in A}(x_i - \mu_A) + m(\mu_A - \mu)^2 \\
&= m\bigl(\sigma_A^2 + (\mu_A - \mu)^2\bigr),
\end{align*}
where the cross-term sum vanishes because $\sum_{i \in A}(x_i - \mu_A) = 0$ by definition of $\mu_A$, and $\sum_{i \in A}(x_i - \mu_A)^2 = m\,\sigma_A^2$ by definition of $\sigma_A^2$.
The removed-channel sum expands as
\begin{equation*}
\sum_{j \in R}(c_j - \mu)^2 = \sum_{j \in R} c_j^2 - 2\mu \sum_{j \in R} c_j + K\mu^2 = Q - 2\mu S + K\mu^2,
\end{equation*}
depending on the constants only through $S$ and $Q$, so no per-channel record of the $c_j$ is required.
Substituting and dividing by $C$ yields:
\begin{equation}
\mu = \frac{m\,\mu_A + S}{m + K}, \qquad
\sigma^2 = \frac{m\bigl(\sigma_A^2 + (\mu_A - \mu)^2\bigr) + \bigl(Q - 2\mu S + K\mu^2\bigr)}{m + K}.
\label{eq:cln}
\end{equation}
The output is then $\hat{x}\,\gamma_{\text{kept}} + \beta_{\text{kept}}$ with $\hat{x} = (x - \mu)/\sqrt{\sigma^2 + \varepsilon}$.
The reconstruction is exact, so every surviving channel receives the same output that standard LayerNorm would have produced on the unreduced $C$-channel input.

When a CompensatedLayerNorm is itself reduced again in a later cycle, the new removed constants are accumulated into $(K, S, Q)$ rather than starting from scratch, so the stored scalars always describe all channels removed across all cycles. 
The storage overhead is three scalars per LayerNorm regardless of how many channels are removed. 

The name "compensated" reflects what the module does: it stores sufficient statistics of the removed channels and uses them to compensate for their absence in the LayerNorm computation, recovering the full-width output on the channels that remain.

\subsection{The Squeeze-Release cycle}
\label{sec:method-cycle}

\begin{figure}[t]
\centering
\includegraphics[width=\textwidth]{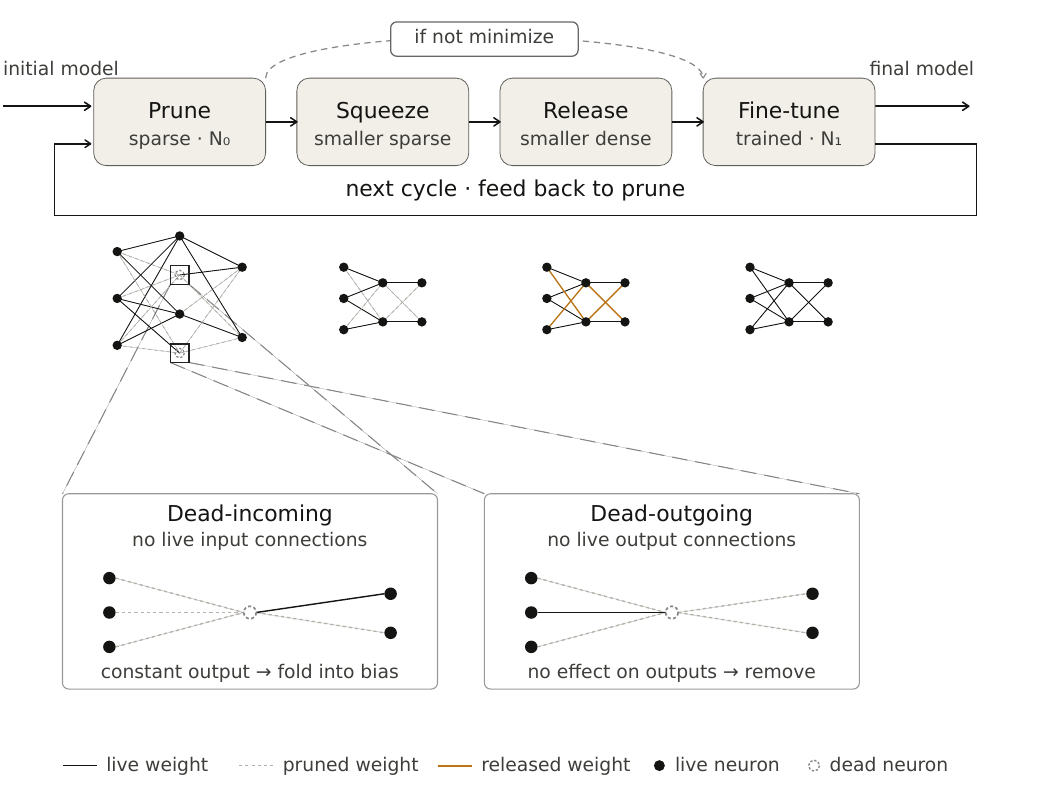}
\caption{The Squeeze-Release cycle. 
Squeeze is a function-preserving structural rewrite that removes dead-incoming and dead-outgoing units. 
Release restores trainable capacity at zero size cost. 
Iteration finds structural redundancy that single-pass minimization cannot.
Cycle is stopped when termination conditions are achieved - either the network size does not decrease, or the max cycle limit has been reached.  
}
\label{fig:cycle}
\end{figure}

We embed the minimization rewrite inside an iterative loop that alternates four steps until the network can no longer be reduced. 
The cycle is illustrated in \Cref{fig:cycle}.

\begin{enumerate}
\item \textbf{Prune.} Starting from the dense network of the previous cycle (or the pretrained model, in the first cycle), apply the pruning step of \Cref{sec:method-setup} with a scheduled sparsity target. 
Fine-tuning passes are interleaved between successive pruning steps so the network can accommodate the increasing sparsity~\cite{zhu2025fggp}. 
Pruning within a cycle is rolled back to the previous epoch if validation accuracy falls below an absolute threshold or drops by more than 10 percentage points in a single epoch.
If the first pruning step of the cycle hits aforementioned conditions, the loop terminates: the network cannot
tolerate further sparsification. 
Otherwise, the output is a masked network in which the lowest-scoring weights are zeroed.

\item \textbf{Squeeze.} Apply the structural rewrite of \Cref{sec:method-minimize}. 
The result is a smaller dense network with no mask buffers, whose forward function matches the pruned predecessor up to floating-point rounding.

\item \textbf{Release.} For every weight that is exactly zero inside the compacted dense tensors of step 2, sample a replacement value from $\mathcal{N}(\mu, \sigma^2) \cdot 0.01$, with $\mu$ and $\sigma$ estimated from the surviving non-zero weights, pooled per-column for FC and over the whole layer for ConvNeXt. 
The small scale is chosen empirically to reintroduce capacity while only mildly perturbing the trained model, in an attempt to encourage further exploration during fine-tuning. 
Tensor shapes and the parameter count are unchanged; only the values at exact-zero positions are modified. 
We note that the choice of distribution is not critical: reinitializing these positions to zero gave comparable results in our tests, indicating that what matters is returning the freed capacity to the network rather than how the released weights are initialized.

\item \textbf{Fine-tune.} Train the released network on the training data for a fixed epoch budget.
\end{enumerate}
The loop terminates when the network can no longer be minimized or a maximum cycle count is reached.

As a comparison baseline we also run a reduced form of the loop in which steps~2 and~3 are omitted: each cycle consists of pruning and fine-tuning only, and the same binary mask is used throughout the run, accumulating zeros across cycles. 
The termination conditions are unchanged.

\subsection{Why release matters: reclaiming wasted capacity}
\label{sec:method-release}

Squeeze produces a smaller dense network, but it does not touch every zero left by pruning. 
The rewrite drops a unit only when an entire row, column, or filter is zero. Zeros that sit inside an otherwise alive row, column, or filter cannot be collapsed: they stay in the compacted dense tensors and are read, multiplied, and accumulated on every forward pass, paying the compute and memory cost of a regular weight while contributing nothing to the output. We refer to these as wasted capacity.

Release reclaims that capacity by replacing the exact zeros with small Gaussian samples, turning frozen positions back into trainable parameters at no change to tensor shape or parameter count. 
After the fine-tune step the released positions hold values shaped by gradient descent rather than a remnant of the previous pruning round. 
The next prune cycle therefore operates on a network whose computation has had the chance to redistribute along previously frozen directions, and tends to find structural redundancy that prune-and-shrink alone cannot.

This argument contradicts the assumption, common in iterative-pruning work, that pruned weights do not meaningfully revive once removed during training~\cite{lym2019prunetrain}. 
It also contrasts with dense-sparse-dense procedures~\cite{han2017dsd,peste2021ac} that re-enable zeros inside a fixed-shape network: those recover trainable capacity but do not change the deployed model's size. 
Squeeze-Release does both, and the two interact, since shrinking lets release operate on a different network each cycle.

\section{Experiments}
\label{sec:experiments}

We evaluate Squeeze-Release on two settings: a fully-connected network on MNIST~\cite{lecun1998mnist} and ConvNeXt-Tiny~\cite{liu2022convnet} on CIFAR-10~\cite{krizhevsky2009cifar}. 
As a separate proof of concept we apply post-hoc minimization to a pruned ViT-Tiny~\cite{dosovitskiy2021vit} checkpoint on ImageNet-1k~\cite{russakovsky2015imagenet} to confirm that the minimization rewrite can be extended to transformer architectures. 
The importance score used for pruning is defined in \Cref{sec:method-setup}.
The No-minimize baseline registers a single binary mask once and accumulates zeros across cycles, in line with standard iterative-pruning protocol; the deployable size of a No-minimize run is obtained by applying the structural rewrite of \cref{sec:method-minimize} once, post hoc, to the final masked model with no additional training. 
Termination follows the cycle protocol of \cref{sec:method-cycle}; the validation-accuracy stopping threshold is specified per setting below.

For the FC setting the model is a fully-connected network of widths [784, 128, 256, 128, 128, 64, 10] with BatchNorm and SELU, totalling $193{,}226$ parameters of which $191{,}104$ are prunable linear weights. 
We pre-train from scratch for $160$ epochs with SGD ($\text{lr}{=}0.1$, momentum $0.9$, weight decay $5{\times}10^{-4}$, cosine schedule, batch size $128$) to set a baseline with accuracy of $98.24\%$. 
We split the standard $60$k training set into $55$k for training and $5$k for validation, report accuracy on the $10$k test set.
Pruning cycles reuse the SGD configuration with a flat learning rate for $160$ epochs each; the between-cycle and post-loop fine-tune runs $20$ epochs at $\text{lr}{=}0.01$ with a cosine schedule, and use a stopping threshold of $80\%$ accuracy on a validation set. 
The sparsity kept-weight ratio follows the cubic schedule~\cite{zhu2017prune}: writing $p \in [0, 1]$ for the fraction of a cycle's pruning epochs elapsed, the target kept ratio is $r(p) = r_f + (r_0 - r_f)\,(1 - p)^3$, decreasing from full density $r_0 = 1$ to final $r_f = 0.002$.
  
We run two configurations with $5$ seeds each: the No-minimize baseline and Squeeze-Release.

For the ConvNeXt-Tiny setting we start from the publicly released checkpoint fine-tuned on CIFAR-10~\cite{javid2024convnextcifar10}, with $27{,}827{,}818$ parameters. 
Inputs are resized to $224{\times}224$ and augmented during training with random $224$-crops (padding $28$) and horizontal flips. 
Pruning cycles use AdamW ($\text{lr}{=}10^{-4}$, weight decay $10^{-2}$, flat schedule, batch size $64$) for $100$ epochs each; between-cycle and final fine-tunes run for $20$ epochs at $\text{lr}{=}5{\times}10^{-5}$ with a cosine schedule. 
The sparsity target follows the same cubic schedule as FC. 
We use a $45$k/$5$k train/validation split of the $50$k training set, report accuracy on the $10$k test set,
and use the same $80\%$ validation stopping threshold. 
Pruning is filter-level on the depthwise convolutions and element-wise on the pointwise (Linear) layers and the classifier. 
We run three configurations with $5$ seeds each: the No-minimize baseline; Squeeze-Release in \textbf{Mode~A}, which uses Stage~A only (\cref{sec:method-convnext}); and Squeeze-Release in \textbf{Mode~AB}, which additionally applies Stage~B residual-stream reduction via \emph{CompensatedLayerNorm} (\cref{sec:method-cln}). 
Stage~C (\cref{sec:method-convnext}) is applied in both modes.

For the ViT proof of concept we use \texttt{vit\_small\_patch16\_224} from \texttt{timm}~\cite{wightman2019timm} pretrained on ImageNet-1k, with $22{,}050{,}664$ parameters across $12$ blocks (embedding width $D{=}384$, $6$ attention heads, MLP inner width $1536$). 
Pruning cycles use AdamW ($\text{lr}{=}5{\times}10^{-5}$, weight decay $5{\times}10^{-2}$, batch size $512$) with two training epochs per pruning update and a $10$-epoch fine-tune between cycles; we sample a $40\%$ random subset of the training set per epoch to fit the wall-clock budget, and the sparsity target follows a linear taper. 
The stopping threshold is set to $60\%$ validation accuracy, which was deemed a reasonable level for more complex ImageNet-1k task evaluation without a resource-intensive search for high-performance training hyper-parameters settings capable to recover network after pruning stages. 
To preserve the multi-head attention shape, $Q$, $K$, $V$ rows at the same output position are scored and masked jointly. 
Minimization follows ConvNeXt Mode~A in logic - Stage~A on the MLP inner dimension and Stage~C on whole blocks - \cref{sec:method-convnext}, and additionally removes fully-zero attention heads. 
We run only the No-minimize baseline followed by post-hoc minimization to illustrate that there are no technical limitations to apply the method to transformer-based networks.

For each run we report final test accuracy after the post-loop fine-tune, the deployable parameter count of the resulting dense network, the mask-alive count $\|m\|_0$ in the last successful cycle and the number of completed cycles.

\section{Results}
\label{sec:results}

\subsection{FC on MNIST}
\label{sec:res-fc}

The pre-trained reference reaches $98.24\%$ test accuracy on the $191{,}104$ prunable Linear weights of the FC model.
\Cref{tab:fc-results} reports the $5$-seed mean and standard deviation for the No-minimize baseline and Squeeze-Release.

Applying the structural rewrite of \cref{sec:method-minimize} once, post hoc, to the No-minimize baseline reduces the model from $191{,}104$ to $18{,}126$ parameters on average, a $10.5\times$ reduction available to any unstructured pruner without changing the training loop. 
Squeeze-Release pushes this further to $4{,}878$ parameters, $3.7\times$ smaller than the post-hoc-minimized baseline or $39.2\times$ smaller than no-minimize baseline. 
The cost is $\sim\!2.8\times$ required cycles compared to the baseline.
The accuracy difference is not statistically significant (Welch's t-test, $p = 0.055$).

The two configurations reach qualitatively different points on the mask-alive vs.\ deployable axis. 
The No-minimize baseline ends with $3{,}354$ non-zero mask entries on average - the metric most unstructured-pruning work reports use - yet the deployable dense network it produces is more than $5\times$ larger than that count would suggest. 
Squeeze-Release ends with more non-zero mask entries than the baseline ($4{,}878$ vs.\ $3{,}354$) but a strictly smaller deployable network: by construction, the mask-alive and deployable counts become equal after each release step.

\begin{table}[ht]
\centering
\small
\begin{tabular}{lrrrr}
\toprule
Configuration & Accuracy (\%) & Mask-alive & Deployable & Cycles \\
\midrule
No-minimize baseline & $95.99 \pm 0.84$ & $3{,}354 \pm 196$ & $18{,}126 \pm 2{,}824$ & $4.2 \pm 1.8$ \\
Squeeze-Release      & $97.03 \pm 0.56$ & $4{,}878 \pm 1{,}584$ & $4{,}878 \pm 1{,}584$ & $11.6 \pm 3.7$ \\
\bottomrule
\end{tabular}
\caption{FC on MNIST: 5-seed mean $\pm$ standard deviation. 
\emph{Mask-alive} is the count of non-zero entries in the persisted pruning mask at the end of the last successful cycle; \emph{Deployable} is the parameter count of the dense network after the structural rewrite of \cref{sec:method-minimize}.
For Squeeze-Release these two values coincide by construction as models use all available capacity.}
\label{tab:fc-results}
\end{table}

\subsection{ConvNeXt-Tiny on CIFAR-10}
\label{sec:res-convnext}

\Cref{tab:convnext-results} reports 5-seed mean and standard deviation for the No-minimize baseline and Squeeze-Release in Modes~A and~AB on the $27{,}827{,}818$-parameter ConvNeXt-Tiny model.

Applying the Mode~AB structural rewrite once, post hoc, to the No-minimize baseline reduces the deployable size to $10.0$M parameters on average, a $2.78\times$ reduction from the full model with no training-loop change. 
Squeeze-Release Mode~AB pushes this to $1.88$M parameters, $5.3\times$ smaller than the post-hoc-minimized baseline and $14.8\times$ smaller than the full pre-trained model, at $90.27\%$ test accuracy versus $90.81\%$ for the baseline. 
Mode~A is less aggressive on size ($2.45$M parameters, $4.1\times$ smaller than the baseline or $11.4\times$ smaller than the full pre-trained model) and reaches $91.14\%$ test accuracy.
No accuracy difference relative to the baseline is statistically significant under Welch's t-test with Bonferroni correction ($p_{adj} = 0.52$ for Mode A, $0.11$ for Mode AB) or raw Welch's t-test ($p = 0.26$ for Mode A, $0.055$ for Mode AB).

The contribution of Stage~B is directly visible from the difference between Mode~A and Mode~AB. 
Enabling residual-stream reduction via \emph{CompensatedLayerNorm} (\cref{sec:method-cln}) shrinks the deployable model by an additional $1.3\times$ on top of Stage~A, providing empirical justification for the LayerNorm rewrite.

A per-call unit test confirms the rewrite stays exact throughout: across all ConvNeXt minimization calls the smaller dense network reproduces the original forward function to within floating-point precision, with maximum absolute logit difference $1.06{\times}10^{-6}$ across all experiments.

The mask-vs-deployable gap behaves qualitatively as on FC but is wider. 
The No-minimize baseline ends with $884$k non-zero mask entries on average while the deployable dense network has $10.0$M parameters: the deployable size is $11.3\times$ larger than the mask-alive count, roughly twice the gap observed on FC.

We have capped the number of cycles to 100 to save computational resources.

\begin{table}[ht]
\centering
\small
\begin{tabular}{lrrrr}
\toprule
Configuration & Accuracy (\%) & Mask-alive (k) & Deployable (M) & Cycles \\
\midrule
No-minimize baseline    & $90.81 \pm 0.42$ & $884 \pm 115$ & $10.01 \pm 1.82$ & $3.6 \pm 1.3$ \\
Squeeze-Release Mode A  & $91.14 \pm 0.44$ & $144 \pm 43$  & $2.45 \pm 0.03$  & $ 97.2 \pm 6.3$ \\
Squeeze-Release Mode AB & $90.27 \pm 0.32$ & $114 \pm 10$  & $1.88 \pm 0.02$  & $100$ \\
\bottomrule
\end{tabular}
\caption{ConvNeXt-Tiny on CIFAR-10: 5-seed mean $\pm$ standard deviation. 
Mask-alive counts are reported in thousands (k); deployable counts in millions (M). 
\emph{Mask-alive} is the count of non-zero entries in the persisted pruning mask at the end of the last successful cycle, capped at 100 cycles limit; \emph{Deployable} is the parameter count of the dense network produced by the structural rewrite of \cref{sec:method-minimize} (Mode~AB applied post hoc for the baseline; Modes~A and~AB applied throughout for Squeeze-Release).}
\label{tab:convnext-results}
\end{table}

\subsection{ViT-Small on ImageNet-1k}
\label{sec:res-vit}

The single-seed run completed five full cycles of the No-minimize protocol. T
he sixth cycle validation-accuracy has dropped below the $60\%$ threshold on its fourth pruning step and was rolled back to the preceding step. 
At the rollback state, the persisted mask carries $3.87$M non-zero entries on the $22{,}050{,}664$-parameter model, a sparsity of $82.45\%$.

Applying the structural rewrite of \cref{sec:method-minimize} once, post hoc, to this masked model yields a deployable dense network of $19.40$M parameters, a $12.0\%$ reduction from the full model. 
The rewrite preserves the forward function up to floating-point rounding (maximum absolute logit difference $5.72{\times}10^{-6}$). 

The mask-vs-deployable gap is $5.0\times$ ($19.40$M deployable against $3.87$M mask-alive), close to the FC ratio and well below the $11.3\times$ measured on ConvNeXt. 
The PoC proves that the minimization rewrite extends also to transformer architectures and that the same metric pathology appears at transformer scale.

\section{Discussion}
\label{sec:discussion}

Across the three types of neural networks of \cref{sec:results}, the \emph{No-minimize} mode leaves a substantial gap between the mask-alive count $\|m\|_0$ and the deployable parameter count produced by the structural rewrite of \cref{sec:method-minimize}.
It is $5.4\times$ on FC, $11.3\times$ on ConvNeXt, and $5.0\times$ on ViT. 
Every parameter inside this gap is a zero that unstructured pruning leaves behind in the compacted dense tensors after the rewrite has done what it can. 
These zeros are read, multiplied, and accumulated on every forward pass while contributing nothing to the output; we have called them wasted capacity in \cref{sec:method-release}. 
The mask-alive count answers a different question than ``how big is the deployable model'', and on architectures where filter granularity and residual-stream channel alignment impose tighter structural constraints, the two answers diverge more.

The release step of the Squeeze-Release cycle (\cref{sec:method-release}) is what converts wasted capacity from a measurement issue into a compression mechanism. 
After each squeeze, the exact-zero positions inside the compacted dense tensors are sampled with small noise calibrated to the surviving weight statistics and trained back into useful parameters. 
The next pruning cycle operates on a network whose computation has had the chance to redistribute along those positions, and tends to find structural redundancy that a single pass cannot reach. 
On ConvNeXt-Tiny this takes the deployable size from a $2.78\times$ reduction under post-hoc minimization of the No-minimize baseline to a $14.8\times$ reduction under Squeeze-Release Mode AB at matched accuracy.
The cycle counts in \cref{tab:convnext-results} are a more direct sign that this capacity is reused. 
The No-minimize baseline, which can never reuse the zeros it accumulates, terminates after $3.6 \pm 1.3$ cycles, whereas Squeeze-Release sustains $97.2 \pm 6.3$ cycles in Mode~A and reaches the $100$-cycle cap in Mode~AB at comparable accuracy. 
Release is the only procedural difference between the configurations, so the large gap in how long the loop keeps finding prunable structure indicates that the freed positions are taken up and trained into useful weights rather than left inert.
On ConvNeXt-Tiny this takes the deployable size from a $2.78\times$ reduction under post-hoc minimization of the No-minimize baseline to a $14.8\times$ reduction under Squeeze-Release Mode AB at matched accuracy.

Squeeze-Release offers two practical entry points. 
Mode A+C is almost free in deployment terms. 
The resulting network reuses the same layer types and block structure as the original, with only the inner widths reduced and occasional full blocks removed. 
On ConvNeXt-Tiny it already provides a $4.1\times$ reduction relative to the No-minimize baseline, at $91.14\%$ test accuracy compared with the baseline's $90.81\%$. 
Mode A+B+C additionally introduces CompensatedLayerNorm (\cref{sec:method-cln}), a lightweight drop-in replacement for the standard LayerNorm that enables exact pruning across residual-stream channels; this earns a further $1.3\times$ reduction in deployable size, with test accuracy ($90.27\%$). 
More broadly, the mask-alive count remains the most widely reported metric in unstructured-pruning work, and it captures something useful: how aggressively a method has driven weights to zero. 
Our measurements suggest it does not by itself answer the question downstream users are usually asking, which is how large the deployed model will be. 
Reporting both the mask-alive count and the post-minimization size when presenting unstructured-pruning results would let that distinction be made explicitly.

The wall-clock cost of iteration is roughly $4.5\times$ that of a single-pass baseline on FC/MNIST and $>30\times$ on ConvNeXt-Tiny where we have set a hard limit on the number of cycles. 
For one-time compression of a deployable model this is a modest cost, but it is not negligible and we do not frame it as free.

The transformer setting is reported as a post-hoc minimization proof of concept on a single seed, without the iterated Squeeze-Release loop. 
Ability to reduce deployable size of a pruned/sparse transformer-based network implies that Squeeze-Release approach can be applied to it as well as to other transformer-like neural networks.

\section{Conclusion}
\label{sec:conclusion}

Exact minimization turns the output of any unstructured pruner into a smaller dense deployable model, and CompensatedLayerNorm extends the rewrite to LayerNorm-equipped architectures. 
Reporting the post-minimization size alongside the mask-alive count gives a fairer basis for comparing pruning methods, since it reflects what is actually deployed.

Iterating the four-step cycle (prune, squeeze, release, fine-tune) reaches smaller final architectures than a single prune-and-minimize pass. 
Release reclaims parameter-space capacity in zero positions of the compacted tensors, and subsequent cycles use that capacity to find new structural redundancy.

\section{Acknowledgments}
Roman Denkin acknowledges support from Centre for Interdisciplinary Mathematics (CIM), Uppsala University. 
Ida Åkerholm acknowledges support from the Swedish Research Council through grant 2024-05664. 
The computations/data handling were enabled by resources provided by the National Academic Infrastructure for Supercomputing in Sweden (NAISS) at UPPMAX (Uppsala University) and Alvis (Chalmers University of Technology), partially funded by the Swedish Research Council through grant agreement no. 2022-06725. The authors would like to thank Professor Orcun Goksel for the access to additional computational resources.

\section{Declaration of generative AI and AI-assisted technologies in the manuscript preparation process}
During the preparation of this work the authors used Claude.AI/Anthropic in order to to improve grammar, find synonyms, typographic checking and generating flow diagrams according to authors specifications. 
After using this tool/service, the authors reviewed and edited the content as needed and take full responsibility for the content of the published article.

\bibliographystyle{elsarticle-num}
\bibliography{references}

\end{document}